\documentclass{article}

\PassOptionsToPackage{numbers,compress,sort}{natbib}
\usepackage[dblblindworkshop, final]{neurips_2025}

\workshoptitle{NPGML}

\usepackage[utf8]{inputenc} % allow utf-8 input
\usepackage[T1]{fontenc}    % use 8-bit T1 fonts
\usepackage{hyperref}       % hyperlinks
\usepackage{url}            % simple URL typesetting
\usepackage{booktabs}       % professional-quality tables
\usepackage{amsfonts}       % blackboard math symbols
\usepackage{nicefrac}       % compact symbols for 1/2, etc.
\usepackage{microtype}      % microtypography
\usepackage{xcolor}         % colors
\usepackage{amsmath}
\usepackage{graphicx}
\usepackage{float}

\title{GaLoRA: Parameter-Efficient Graph-Aware LLMs for Node Classification}

\author{%
 Mayur Choudhary \quad Saptarshi Sengupta \quad Katerina Potika \\
  Department of Computer Science, San Jose State University\\
  \texttt{\{mayur.choudhary, saptarshi.sengupta, katerina.potika\}@sjsu.edu}
}

\begin{document}

\maketitle

\begin{abstract}
%With the rapid rise of large language models (LLMs) and their ability to capture semantic relationships, LLMs are being used across a wide range of applications. 
The rapid rise of large language models (LLMs) and their ability to capture semantic relationships has led to their adoption in a wide range of applications.
Text-attributed graphs (TAGs) are a notable example where LLMs can be combined with Graph Neural Networks (GNNs) to improve the performance of node classification. In TAGs, each node is associated with textual content and such graphs are commonly seen in various domains such as social networks, citation graphs, recommendation systems, etc. Effectively learning from TAGs would enable better representations of both structural and textual representations of the graph and improve decision-making in relevant domains. We present GaLoRA, a parameter-efficient framework that integrates structural information into LLMs. GaLoRA demonstrates competitive performance on node classification tasks with TAGs, performing on par with state-of-the-art models with just 0.24\% of the parameter count required by full LLM fine-tuning. We experiment with three real-world datasets to showcase GaLoRA's effectiveness in combining structural and semantical information on TAGs.
\end{abstract}

\section{Introduction}
TAGs present a unique challenge in machine learning by requiring the simultaneous representation of both graph structure and rich textual information \citep{yan2023cstag,jin2024llmgraphs}. Traditional approaches typically leverage GNNs to capture structural dependencies \citep{hamilton2017sage} or Pretrained Language Models (PLMs) to process semantic content \citep{vaswani2017attention}. Joint models that combine both have shown promise but are often computationally expensive and difficult to scale.

We propose GaLoRA (Graph-aware Low-Rank Adaptation), a modular and efficient framework that enables LLMs to incorporate structural information from graphs without requiring joint training. GaLoRA follows a two-phase design: the first in which a GNN learns structure-aware embeddings that are later injected into the LLM during fine-tuning using Low-Rank Adaptation \citep{hu2021lora}. This decoupling reduces training overhead while preserving the benefits of structure-semantic fusion.

Through experiments on various TAG datasets, we demonstrate that GaLoRA achieves performance on par with recent baselines \citep{huang2024graphadapter} while requiring significantly fewer parameters, making it ideal for deployment in resource-constrained real-world settings.

\section{Related Work}

Recent approaches have explored combining GNNs and PLMs to address the challenges of learning on text-attributed graphs. One such method is GLEM \citep{zhao2023glem}, which uses the Expectation-Maximization (EM) framework to train the GNN and LLM in alternating steps. GLEM uses pseudo-labels to iteratively refine both models without requiring joint backpropagation, thereby reducing computational cost. However, its reliance on pseudo-label quality makes it sensitive to noise, and its iterative nature may still be computationally heavy for large-scale graphs.

Another efficient integration method is TAPE \citep{he2024tape}, which leverages LLMs as a service to generate explanations and predictions via prompting, and then uses a smaller language model to convert these explanations into embeddings consumable for GNNs. This modular approach avoids fine-tuning the LLM, significantly reducing computation. However, its strong dependence on manually crafted prompts and the variable interpretability of LLM-generated explanations can lead to inconsistent performance.

More recently, GraphAdapter \citep{huang2024graphadapter} introduced a parameter-efficient strategy that freezes the LLM and incorporates a lightweight GNN adapter to inject graph structure into the LLM's hidden representations. While this approach offers scalability and generalization, it may limit the model’s ability to adapt to task-specific semantic knowledge, as the LLM remains frozen throughout. In contrast to the discussed works, our proposed framework, GaLoRA, aims to preserve modularity and efficiency by decoupling the training of GNN and LLM modules, while still allowing direct integration of structural information into the LLM through LoRA-based adaptation during fine-tuning.

\section{Methodology}
\label{headings}

To provide a modular approach that is efficient, we present GaLoRA, a framework to fine-tune LLMs with graph context using parameter-efficient training. The framework decouples structural and semantic learning; it trains the GNN and LLM modules separately and integrates the learned representation by injecting structural embeddings into the language model during fine-tuning. With the use of this framework, the goal is to fine-tune the language model such that it aligns with both the textual content and the structural information of the TAG. To ensure the language model adaptation is efficient, the framework utilises LoRA \citep{hu2021lora} by adding small trainable low-rank matrices into the frozen layers of the pretrained language model. This enables the LLM to efficiently train and achieve the small delta required for the task-specific adaptation.

The framework operates in two distinct phases. In the first phase, a GNN model is trained on the TAG for the node classification task to extract rich structure-aware node embeddings. In the second phase, the LLM is fine-tuned on a separate node classification task using the node text embeddings as the input. During the fine-tuning, the structure-aware embeddings are injected into selected layers of LLMs, enabling them to incorporate the structural information along with the semantic understanding of the attached text. This modular design decouples structure and language learning while still allowing the LLM to benefit from the structural context.

\subsection{Phase-1: GNN Training}

The first component of the framework is a GNN that models the structural dependencies between nodes. Each node is represented by a feature vector derived from the textual content associated with it. These initial node embeddings are obtained using a LLM encoder, 
identical to the language model employed in the second phase of the framework. In our experiments, we primarily use GraphSAGE \citep{hamilton2017sage} as the GNN, due to its effectiveness in aggregating information from neighboring nodes.

\begin{figure}[ht]
    \hspace{25mm}
    \includegraphics[width=0.6\linewidth]{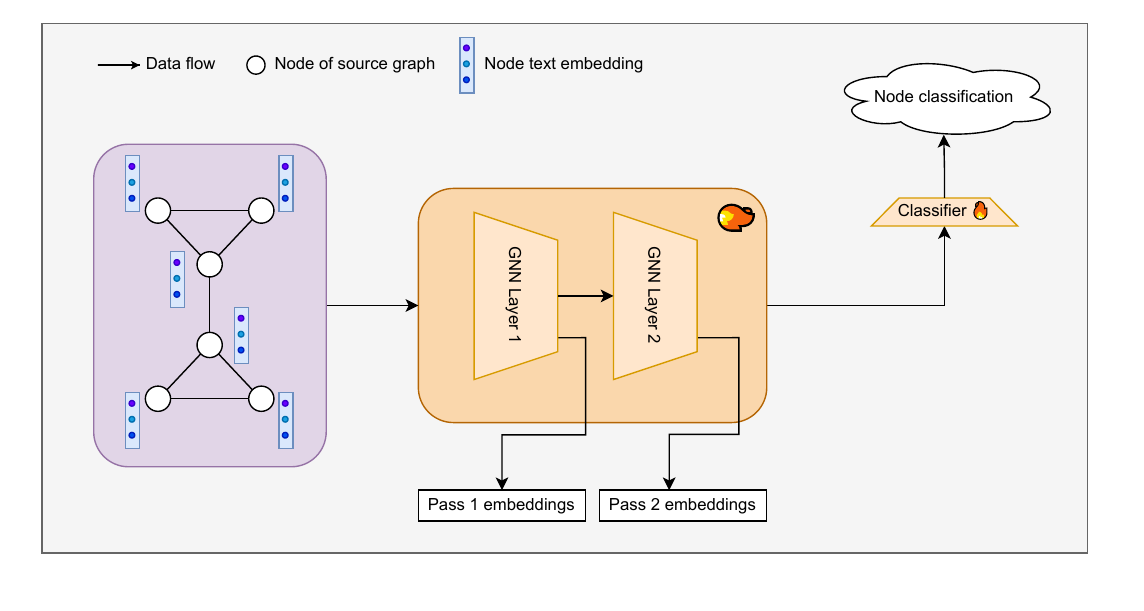} % Adjust width as needed
    \caption{GaLoRA Phase-1 training}
    \label{fig:Phase~1}
\end{figure}

Figure~\ref{fig:Phase~1} shows that the architecture has two message passing layers present in the GNN model, which help capture the 1-hop and 2-hop neighborhood for each node. The first layer output is stored in an intermediate matrix called Pass-1, and the second layer output is similarly stored in the Pass-2 matrix. The information aggregation from the neighbors is done using mean pooling, which is then followed by a non-linear transformation.

\paragraph{\textbf{Pass-1 (1-hop aggregation):}} 
\begin{equation}
H^{(1)}_v = \text{ReLU} \left( W^{(0)} \cdot \text{CONCAT}\Big( X_v, \; \text{MEAN}\{ X_u : u \in \mathcal{N}(v) \} \Big) + b^{(0)} \right)
\end{equation}

\paragraph{\textbf{Pass-2 (2-hop aggregation):}} 
\begin{equation}
H^{(2)}_v = \text{ReLU} \left( W^{(1)} \cdot \text{CONCAT}\Big( H^{(1)}_v, \; \text{MEAN}\{ H^{(1)}_u : u \in \mathcal{N}(v) \} \Big) + b^{(1)} \right)
\end{equation}

\noindent
where:
\begin{itemize}
  \item \( X_v \in \mathbb{R}^d \) is the text embedding of node \(v\),
  \item \( \mathcal{N}(v) \) denotes the set of neighbors of node \(v\),
  \item \( \text{MEAN}\{\cdot\} \) is the neighborhood aggregation function,
  \item \( W^{(0)} \), \( W^{(1)} \) are trainable weight matrices for each layer,
  \item \( b^{(0)} \), \( b^{(1)} \) are learnable bias vectors,
  \item \( \text{ReLU}(\cdot) \) is the non-linear activation function,
  \item \( H^{(1)}_v \) and \( H^{(2)}_v \) are the GraphSAGE embeddings after 1-hop and 2-hop aggregation, respectively.
\end{itemize}

The output embeddings from the Pass-2 layer for each node are passed through a classifier to perform the supervised node classification task. This training objective ensures that the embeddings produced by both Pass-1 and Pass-2 capture meaningful structural information for accurate classification. The classifier used in our research is a lightweight MLP layer.

\subsection{Phase-2: LLM fine-tuning}

\begin{figure} [h]
    \hspace{8mm}
    \includegraphics[width=0.9\linewidth]{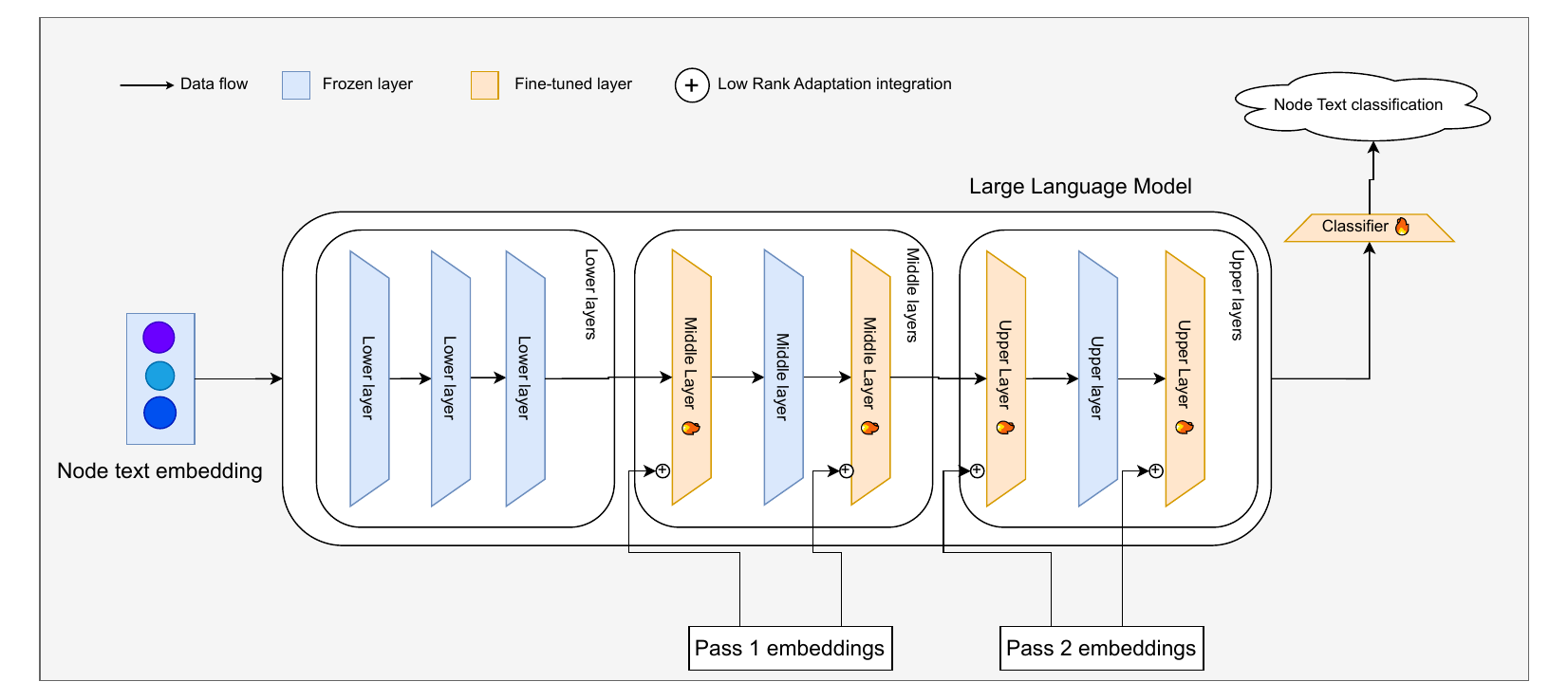} % Adjust width as needed
    \caption{GaLoRA Phase-2 training}
    \label{fig:Phase2}
\end{figure}

In the second phase, the structure-aware node embeddings obtained from GNN are integrated into the LLM to align the eventual output embeddings based on the structural context in addition to the textual content. Instead of fine-tuning the entire LLM, we integrated LoRA only into certain layers of the model. In our setup, we introduce structural information into the LLM by integrating node representations derived from the GNN (both Pass-1 and Pass-2) directly into the middle and upper layers of the language model. The architecture is shown in Figure~\ref{fig:Phase2}.

To integrate the two modalities efficiently, we use a low-rank adaptation mechanism as follows:

\begin{equation}
Z = W_{\text{C}} \cdot \left( \alpha \cdot W_{\text{A}} H_1 + (1 - \alpha) \cdot W_{\text{B}} H_2 \right)
\end{equation}

\noindent
where:
\begin{itemize}
  \item \( H_1 \in \mathbb{R}^{d \times T} \) represents the hidden states from the previous layer of the LLM,
  \item \( H_2 \in \mathbb{R}^{g \times T} \) represents the structural node embeddings from the GNN broadcast across all \(T\) tokens 
  \item \( W_{\text{A}} \in \mathbb{R}^{r \times d} \) projects the LLM input to a lower dimension,
  \item \( W_{\text{B}} \in \mathbb{R}^{r \times g} \) projects the graph embedding to the same low-rank space,
  \item \( W_{\text{C}} \in \mathbb{R}^{d \times r} \) projects the fused representation back to the original dimension,
  \item \( \alpha \in [0, 1] \) is a learnable gate that balances between the text and structure inputs,
  \item \( Z \in \mathbb{R}^{d \times T} \) is the final adapted input passed into the frozen LLM layer.
  \item \( d \): dimensionality of the LLM hidden states,
  \item \( g \): dimensionality of GNN output embeddings (per node),
  \item \( r \): low-rank dimension used for parameter-efficient adaptation,
  \item \( T \): token sequence length for each node's textual content.
\end{itemize}

This formulation allows us to inject structure-aware representations into the
language model in a parameter-efficient manner, without modifying or retraining the
full LLM architecture. As $r << d$ and $r << g$, where d is the dimension of input embeddings and g of graph embeddings, the number of parameters being trained is far less than that during fine-tuning the entire LLM model. The middle layers of LLM are injected with the Pass-1 embeddings and the upper layers with Pass-2 embeddings. This is based on the understanding that middle layers help form context between words, and upper layers help with context formation of higher layers.

The structural embedding from the GNN and the token embeddings from the LLM input layer are each projected into a shared lower-dimensional space of rank  $r$, where $r \ll d$ and $r \ll g$, with $d$ and $g$ being the LLM layer input embedding dimension and GNN module output embedding dimension, respectively. These low-rank projections are then fused via a learnable gate and transformed to the LLM layer output dimension before being passed into the next layer. This injection mechanism enables GaLoRA to efficiently incorporate neighborhood context into the semantic understanding pipeline without retraining the full LLM, significantly reducing computational overhead.

\subsection{Design choices and limitations}

One of the key design choices of the framework is the injection of Pass-1 and Pass-2 embeddings in the middle and upper layers of the LLM encoder in Phase-2, respectively. This allows the LLM to be aware of the immediate neighborhood through Pass-1 embeddings while it forms an understanding of the local context of the tokens. At later stages, having a wider view of the graph through Pass-2 embeddings enables the LLM encoder to reason over a wider context during the final semantic stages.

Another important design element is the use of a learnable gate parameter. In the integration of graph embeddings in the LLM encoder, this helps the LLM regulate the structural influence on LLM for the node classifications. Despite its effectiveness, this framework is currently only evaluated for node classification tasks. 

We use the cross-entropy loss in both phases of GaLoRA because it is the standard choice for node classification tasks and is used by most benchmark models. This allowed us to make fair comparisons with previous work.

We have also adopted stratified splits to balance label distributions, though this may introduce some risk of information leakage in GNNs due to message passing. Future work could mitigate this with benchmark or structure-aware splits. Further work will also explore extending GaLoRA beyond node classification to tasks such as link prediction, graph classification, and advanced fusion designs, as well as validating performance on newly proposed TAG benchmarks and diverse datasets.

\section{Experimentation}

\subsection{Datasets}
We evaluate GaLoRA on three text-attributed graph (TAG) datasets: Instagram, Reddit \citep{huang2024graphadapter_datasets}, and ArXiv \citep{hu2020ogb}. Each dataset contains node-level textual content and edges representing structural relationships. Table~\ref{tab:dataset-summary} summarizes their key statistics.

\textbf{Instagram:} Nodes represent users with bios as text; the task is binary classification of commercial vs.\ non-commercial users.

\textbf{Reddit:} Nodes represent users with text from their last three posts; the task is binary classification of popular vs.\ non-popular users.

\textbf{ArXiv:} Nodes represent papers with titles and abstracts as text; task is $40$-class paper categorization. Due to resource constraints, we evaluate on a $46$K-node subgraph released by prior work.

\begin{table}[H]
\centering
\small
\caption{Summary of datasets used in experiments.}
\begin{tabular}{lcccccc}
\toprule
\textbf{Dataset} & \textbf{\# Nodes} & \textbf{\# Edges} & \textbf{\# Tokens} & \textbf{Split (\%)} & \textbf{\# Classes} & \textbf{Metric} \\
\midrule
ArXiv      & 46,198   &  78,548   & 35,920,710   & 54/18/28     & 40   & Accuracy \\
Instagram  & 11,339    & 144,010     & 579,263      & 80/10/10     & 2    & ROC-AUC \\
Reddit     & 33,434    & 198,448     & 6,748,436    & 80/10/10     & 2    & Accuracy \\
\bottomrule
\end{tabular}
\label{tab:dataset-summary}
\end{table}

\subsection{Experimental Setup}
All experiments were conducted in Google Colab using NVIDIA A100 GPUs (52\, GB VRAM). Implementations utilised PyTorch \citep{paszke2019pytorch}, PyTorch Geometric \citep{fey2019pyg}, and HuggingFace Transformers \citep{wolf2020transformers} libraries, with tokenization handled by the respective pretrained model’s tokenizer.

We used GraphSAGE \citep{hamilton2017sage} with two message passing layers to generate structure-aware node embeddings of size 64. For the language modeling component, we experimented with both GPT-2 \citep{radford2019gpt2} and RoBERTa \citep{liu2019roberta}, applying LoRA-based adaptation to 6 transformer layers (3 middle and 3 upper). The output embeddings from the GNN were aligned with the tokenized text representations and injected into the LLMs during fine-tuning. 

Token lengths varied across datasets (96 for Instagram, 128 for Reddit, and 256 for ArXiv) based on the average node text length. Dataset splits and evaluation metrics are discussed in their respective sections, while training hyperparameters and ablation studies (e.g., LoRA rank) are detailed in the appendix.

\subsection{Results}

This section discusses the results of the method and its performance compared to baseline models. All experiments use GraphSAGE as the GNN. While prior work pairs it with Llama-13B as the pretrained language model, we use smaller LMs (GPT-2 and RoBERTa) due to computational constraints.

\subsubsection{Performance evaluation}

Table~\ref{tab:llm-controlled} reports a \emph{LLM-controlled} comparison where both GaLoRA and GraphAdapter use the same PLM (RoBERTa or GPT-2). Instagram is evaluated with ROC–AUC; Reddit and ArXiv with Accuracy. We use GraphAdapter as the primary baseline, as it reports state-of-the-art results on the TAG benchmarks utilized. GaLoRA is competitive with, and often surpasses, GraphAdapter across datasets and backbones, with the largest margins on ArXiv and Instagram under GPT-2. We use a fixed stratified train/val/test split and five training seeds (0–4), reporting mean and variance across the multiple runs. For context beyond the LLM-controlled setting, Appendix Table~\ref{tab:results-general} discusses the results reported by prior work using larger LLMs (e.g., LLaMA-13B \citep{touvron2023llama2} for GraphAdapter and DeBERTa-Large \citep{he2021deberta} for GLEM) alongside GaLoRA (GPT-2).

\begin{table}[H]
\centering
\caption{LLM-controlled comparison (same LLM across methods). Instagram uses ROC–AUC; Reddit/ArXiv use Accuracy. For GaLoRA, results are mean ± std over 5 seeds (0–4). Baseline results (GNN, GraphAdapter) are taken directly from their reported values.}
\small
\setlength{\tabcolsep}{4pt} % adjust column spacing if needed
\resizebox{\textwidth}{!}{%
\begin{tabular}{lcccccc}
\toprule
 & \multicolumn{2}{c}{\textbf{ArXiv}} & \multicolumn{2}{c}{\textbf{Instagram}} & \multicolumn{2}{c}{\textbf{Reddit}} \\
\cmidrule(lr){2-3}\cmidrule(lr){4-5}\cmidrule(lr){6-7}
\textbf{Model} & RoBERTa & GPT-2 & RoBERTa & GPT-2 & RoBERTa & GPT-2 \\
\midrule
GNN (PLM)$^{\dagger}$      & 0.7129 {\scriptsize(0.0013)} & 0.7174 {\scriptsize(0.0019)} & 0.6123 {\scriptsize(0.0063)} & 0.6019 {\scriptsize(0.0124)} & 0.6191 {\scriptsize(0.0043)} & 0.6282 {\scriptsize(0.0036)} \\
GraphAdapter$^{\dagger}$   & \textbf{0.7273} {\scriptsize(\textbf{0.0021})} & 0.7325 {\scriptsize(0.0022)} & 0.6292 {\scriptsize(0.0033)} & 0.6276 {\scriptsize(0.0034)} & 0.6379 {\scriptsize(0.0061)} & 0.6441 {\scriptsize(0.0022)} \\
\textbf{GaLoRA (Ours)}     & 0.7234 {\scriptsize(0.0014)} & \textbf{0.7550} {\scriptsize(\textbf{0.0048})} & \textbf{0.6392} {\scriptsize(\textbf{0.0093})} & \textbf{0.6420} {\scriptsize(\textbf{0.0046})} & \textbf{0.6464} {\scriptsize(\textbf{0.0035})} & \textbf{0.6611} {\scriptsize(\textbf{0.0049})} \\
\bottomrule
\end{tabular}
}

\vspace{0.25em}
\footnotesize $^{\dagger}$ Baseline numbers are quoted from the GraphAdapter \citep{huang2024graphadapter} paper; we did not retrain these models. \emph{ArXiv:} Results use the 46k-node subgraph provided by GraphAdapter to keep experiments feasible on our hardware.
\label{tab:llm-controlled}
\end{table}

\subsubsection{Parameter Efficiency Comparison}
Table~\ref{tab:param-comp} compares pretrained language models (PLMs) and the number of trainable parameters relative to each method’s own backbone size. GaLoRA trains only $0.18$M parameters in the GNN (Phase~1) and $0.115$M  for the LoRA layers in the LLM (Phase~2), for a total of $0.295$M trainable parameters (\textbf{0.238\%} of GPT-2). In contrast, GLEM fine-tunes the entire DeBERTa-Large model, while GraphAdapter trains only a GNN and fusion layer (\textbf{0.015\%} of LLaMA-13B) but does not fine-tune the LLM to capture semantic knowledge from the textual content. GaLoRA combines low-rank adaptation with structural embeddings, achieving competitive performance with the smallest trainable parameter footprint among methods that perform semantic fine-tuning. Note that the percentage of GaLoRA is reported with respect to GPT-2; when applied to larger models such as LLaMA-13B, the fraction of trainable parameters would be even smaller, since only a subset of layers is adapted and GNN training remains unchanged.

\begin{table}[h]
\centering
\small
\caption{Comparison of PLMs and trainable parameters. Relative \% is computed w.r.t.\ each method’s own PLM parameter count. Phase~1: GNN module; Phase~2: LoRA-injected LLM layers.}
\label{tab:param-comp}
\begin{tabular}{l l r r r}
\toprule
Model & PLM & \#~PLM Params & \#~Trainable Params & Relative to own PLM \\
\midrule
GLEM & DeBERTa-Large & 435M & 435M & 100\% \\
GraphAdapter & LLaMA~2--13B & 13B & 2M & 0.015\% \\
\textbf{GaLoRA (Ours)} & GPT-2 & 124M & 0.295M & \textbf{0.238\%} \\
\bottomrule
\end{tabular}
\end{table}

\paragraph{Summary:}
GaLoRA delivers competitive accuracy to state-of-the-art structural modeling approaches while training less than one percent of its LLM’s parameters. Its two-phase design enables structural and semantic adaptation with minimal computational cost, making it a practical and scalable solution for resource-constrained settings.

\section{Conclusion}

In this work, we introduced GaLoRA (Graph-aware Low-Rank Adaptation), a modular and parameter-efficient framework for enhancing LLM performance on text-attributed graphs. By decoupling structural and semantic learning into two phases and injecting GNN-derived structural embeddings during fine-tuning, GaLoRA achieves strong classification results while significantly reducing training overhead. Our experiments demonstrate that even smaller language models like GPT-2 benefit meaningfully from structural context, highlighting GaLoRA’s potential in resource-constrained settings. The framework’s modular design also opens up future extensions to other graph tasks, such as link prediction or clustering, and potential exploration of richer GNN backbones or fusion strategies. Overall, GaLoRA offers a promising direction for scalable, structure-aware language model adaptation.

\newpage

\bibliographystyle{unsrtnat}
\bibliography{references}

\newpage

%%%%%%%%%%%%%%%%%%%%%%%%%%%%%%%%%%%%%%%%%%%%%%%%%%%%%%%%%%%%

\appendix

\section{Technical Appendices and Supplementary Material}

\subsection{LoRA Integration Details}

We applied LoRA modules to both GPT-2 and RoBERTa backbones, each consisting of 12 transformer layers. To incorporate structural information, we added adapters to six layers per model. Pass-1 embeddings were injected into the middle layers, while Pass-2 embeddings were injected into the upper layers (Table~\ref{tab:lora-layers}). 

\begin{table}[h]
\centering
\caption{LoRA layer selection for GPT-2 and RoBERTa backbones.}
\label{tab:lora-layers}
\begin{tabular}{lcc}
\toprule
Model & Pass-1 embedding & Pass-2 embedding \\
\midrule
GPT-2 (12 layers)     & 5, 6, 7 & 9, 10, 11 \\
RoBERTa (12 layers)   & 5, 6, 7 & 9, 10, 11 \\
\bottomrule
\end{tabular}
\end{table}

LoRA was applied to different projection weights in each backbone (Table~\ref{tab:lora-config}). For GPT-2, we modified the \texttt{c\_attn} and \texttt{c\_proj} projections. For RoBERTa, LoRA was added to the query, key, value, and dense layers of self-attention. We experimented with ranks $r \in \{2, 4, 8\}$, with $r=4$ serving as the default. LoRA weights were initialized with small random values so that the frozen backbone dominated at the start of training.

\begin{table}[h]
\centering
\caption{LoRA configuration for GPT-2 and RoBERTa.}
\label{tab:lora-config}
\begin{tabular}{lccc}
\toprule
Model & Adapted Weights  \\
\midrule
GPT-2     & c\_attn, c\_proj  \\
RoBERTa   & q, k, v, dense \\
\bottomrule
\end{tabular}
\end{table}

Table~\ref{tab:lora-efficiency} reports parameter efficiency for GPT-2 and RoBERTa with rank $r=4$. 
For GPT-2, the number of trainable parameters scales approximately linearly with $r$ (e.g., $\sim$58K at $r=2$ and 230K at $r=8$), 
while the backbone remains fixed at 124.6M parameters. 
Even at the highest tested rank, fewer than 0.2\% of backbone parameters were updated, keeping total trainable parameters well below one million.

\begin{table}[h]
\centering
\caption{Trainable parameter footprint for GPT-2 and RoBERTa with rank = 4. }
\label{tab:lora-efficiency}
\begin{tabular}{lccc}
\toprule
Model   & Total Params & Trainable Params & Relative (\%) \\
\midrule
GPT-2   & 124.6M & 115.2K & 0.09\% \\
RoBERTa & 125.0M & 175.1K & 0.14\% \\
\bottomrule
\end{tabular}
\end{table}

LoRA training used cross-entropy loss with AdamW. Default hyperparameters are summarized in Table~\ref{tab:lora-training}. In practice, some hyperparameters such as weight decay were tuned per dataset to stabilize training and improve generalization. A scheduler was implemented but not used in final runs. All LoRA adapters were implemented manually in PyTorch without relying on the PEFT library, which allowed direct integration with HuggingFace transformer layers.

\begin{table}[H]
\centering
\caption{Training configuration for LoRA adapters (default)}
\label{tab:lora-training}
\begin{tabular}{ll}
\toprule
Hyperparameter & Value \\
\midrule
Loss function     & Cross-Entropy \\
Optimizer         & AdamW \\
Learning rate     & $3 \times 10^{-4}$ \\
Weight decay      & $10^{-2}$ \\
Batch size        & 32 \\
Implementation    & Manual (PyTorch, HuggingFace) \\
\bottomrule
\end{tabular}
\end{table}

\subsection{LoRA rank study}
Table~\ref{tab:rank-study} reports ROC-AUC scores for the Instagram dataset with different LoRA ranks. Performance increased from ($r=2$) to ($r=4$), and reached its highest value at ($r=8$).  However, the gain over ($r=4$) was marginal relative to the additional parameters introduced, suggesting that moderate ranks provide the best trade-off between performance and parameter efficiency.

\begin{table}[h]
\centering
\small
\caption{Effect of LoRA rank on Instagram performance.}
\label{tab:rank-study}
\begin{tabular}{lc}
\toprule
Rank ($r$) & ROC-AUC \\
\midrule
2 & 0.6347 \\
4 & 0.6420 \\
8 & 0.6421 \\
\bottomrule
\end{tabular}
\end{table}

\subsection{Prompt Engineering Analysis}
We also explored the effect of lightweight prompt engineering on performance. 
The setup used GraphSAGE with GPT-2 on the Instagram dataset, where different textual prompts were prepended to the node text before classification. 
Table~\ref{tab:prompt-analysis} reports the ROC-AUC scores for different prompt variations. 

\begin{table}[h]
\centering
\small
\caption{Effect of prompt engineering on Instagram dataset performance.}
\label{tab:prompt-analysis}
\begin{tabular}{lc}
\toprule
Prompt & ROC-AUC \\
\midrule
No prompt (raw text) & 0.6283 \\
\texttt{Classify:} & 0.6305 \\
\texttt{Classify this instagram account bio:} & 0.6344 \\
\texttt{Classify instagram account is commercial or not:} & 0.6420 \\
\bottomrule
\end{tabular}
\end{table}

Even minimal prompting led to considerable improvements over the no-prompt baseline. The most detailed instruction achieved the highest performance, suggesting that explicit task guidance can help align the model to the classification objective, although with relatively modest gains.

\subsection{Comparison of GaLoRA Configurations}

Table~\ref{tab:config-comparison} compares the performance of different GaLoRA model configurations across datasets. Instagram results are reported using ROC-AUC, while Reddit and ArXiv results use accuracy.

\begin{table}[h]
\centering
\caption{Performance comparison of GaLoRA model variants across datasets.}
\label{tab:config-comparison}
\begin{tabular}{lccc}
\toprule
\textbf{Model} & \textbf{Instagram} & \textbf{Reddit} & \textbf{ArXiv} \\
\midrule
GraphSAGE + GPT-2    & 0.6420 & 0.6611 & 0.7550 \\
GraphSAGE + RoBERTa  & 0.6392 & 0.6464 & 0.7234 \\
GAT + GPT-2          & 0.6616 & 0.6613 & 0.7569 \\
\bottomrule
\end{tabular}
\end{table}

\subsection{Comparison with Prior Work (Different Base LMs)}
\label{app:crosspaper}

Table~\ref{tab:results-general} reports ROC-AUC for Instagram and Accuracy for Reddit and ArXiv. Best results are highlighted in bold.

As described in Table~\ref{tab:results-general} GaLoRA outperforms all baseline models for the Reddit dataset, achieving the highest accuracy score across the models. For the Instagram and ArXiv datasets, it underperforms when compared to the GraphAdapter model, although it is to be noted that GraphAdapter utilised Llama-13B as its LLM compared to GPT-2 of GaLoRA. These results show that GaLoRA is capable of integrating structural awareness to LLM-based predictions in a parameter-efficient and effective manner.

\begin{table}[H]
\centering
\caption{Performance comparison across datasets. }
\begin{tabular}{lccc}
\toprule
\textbf{Model} & \textbf{ArXiv} & \textbf{Instagram} & \textbf{Reddit} \\
\midrule
GNN-only              & 0.6980 & - & - \\
LLM-only              & 0.7541 & 0.6248 & 0.6123 \\
LLM (LoRA)            & 0.7454 & 0.5910 & 0.5740 \\
GLEM                       & 0.7315 & 0.6105 & 0.6221 \\
TAPE                            & 0.7672 & - & - \\
GraphAdapter          & \textbf{0.7707} & \textbf{0.6513} & 0.6461 \\
\textbf{GaLoRA (Ours)}                            & 0.7550 & 0.6420 & \textbf{0.6611} \\
\bottomrule
\end{tabular}
\label{tab:results-general}
\end{table}

{\footnotesize \textit{Note.} All structural modeling approaches use GraphSAGE as the GNN. 
GLEM and GraphAdapter employ DeBERTa-Large and LLaMA-13B, respectively. 
GaLoRA uses GPT-2 with LoRA-based structural embedding injection. 
LLM-only and LLM (LoRA) baselines use textual features only. 
A subgraph of ArXiv was used for GaLoRA due to computational constraints.}

\end{document}